\documentclass[10pt]{article}

\usepackage{amsmath}
\usepackage{amssymb}

\usepackage{graphicx}

\usepackage{cite}

\usepackage{color} 

\usepackage[labelfont=bf,labelsep=period,justification=raggedright]{caption}

\makeatletter
\renewcommand{\@biblabel}[1]{\quad#1.}
\makeatother

\date{}

\begin{document}

\begin{flushleft}
{\Large
\textbf{Latent semantics of action verbs reflect phonetic parameters of intensity and emotional content}
}\newline 
\\
Michael~Kai~Petersen$^{\ast}$ \newline
\\
\bf Cognitive Systems \newline DTU Compute, Building 324 \newline Technical University of Denmark \newline DK-2800 Kgs.Lyngby, Denmark
\\
$\ast$ mkai@dtu.dk
\end{flushleft}

\section*{Abstract}

Conjuring up our thoughts, language reflects statistical patterns of word co-occurrences which in turn come to describe how we perceive the world. Whether counting how frequently nouns and verbs combine in Google search queries, or extracting eigenvectors from term document matrices made up of Wikipedia lines and Shakespeare plots, the resulting latent semantics capture not only the associative links which form concepts, but also spatial dimensions embedded within the surface structure of language. As both the shape and movements of objects have been found to be associated with phonetic contrasts already in toddlers, this study explores whether articulatory and acoustic parameters may likewise differentiate the latent semantics of action verbs. Selecting 3 $\times$ 20 emotion, face, and hand related verbs known to activate premotor areas in the brain, their mutual cosine similarities were computed using latent semantic analysis LSA, and the resulting adjacency matrices were compared based on two different large scale text corpora; HAWIK and TASA. Applying  hierarchical  clustering to identify common structures across the two text corpora, the verbs largely divide into combined mouth and hand movements versus emotional expressions. Transforming the verbs into their constituent phonemes, and projecting them into an articulatory space framed by tongue height and formant frequencies, the clustered small and large size movements appear differentiated by front versus back vowels corresponding to increasing levels of arousal. Whereas the clustered emotional verbs seem characterized by sequences of close versus open jaw produced phonemes, generating up- or downwards shifts in formant frequencies that may influence their perceived valence. Suggesting, that the latent semantics of action verbs reflect parameters of intensity and emotional polarity that appear correlated with the articulatory contrasts and acoustic characteristics of phonemes.\newline

\section*{Introduction}

If language adapted to the brain like a virus \cite{Deacon:1998vn}, constrained by sensorimotor circuits linking articulatory gestures with aspects of motion \cite{Glenberg:2011vn}, then parameters of size and intensity might also potentially be reflected in the latent semantics of words. Spatiotemporal metaphors are ubiquitous in phrases like `hitting the road', `christmas is approaching' or `thinking out of the box',  where we reinterpret ourselves as objects that are subject to forces of gravity or moving along virtual time lines \cite{Lakoff:1999fv}. Aspects of motion appear to constrain how we internally represent affordances for potential action,  as perceptual states  are reenacted from memory traces formed by sensorimotor circuits \cite{Engel:2013aa}. Likewise, emotions are metaphorically constrained by spatial parameters; we process positive and negative words faster depending on whether they are presented at the top or bottom of a screen, as upwards or downwards is perceived as vertically congruent with pleasant and unpleasant connotations respectively \cite{Meier:2004aa}. Adding to a growing amount of evidence for embodied cognition \cite{Barsalou:2008kx}, where not only action verbs like `push' are associated with trajectories, but also terms like `argue' and `respect' appear to be grounded in a conceptual space framed by horizontal and vertical axes \cite{Barsalou:2008vn}. Recent studies have demonstrated that spatial dimensions can be retrieved from the surface structure of sentences describing horizontal and vertical movements \cite{Louwerse:2011aa}. Similarly that it is feasible to geographically map out the relative distances between cities based on how they as words co-occur in news articles \cite{Louwerse:2009aa} or in fiction like ``Lord of the Rings'' \cite{Louwerse:2012aa}. Underlying parameters of size and intensity may even be reflected in the phonetic building blocks of language, as behavioral studies have shown that high front vowels are perceived as lighter and associated with smaller organisms than words involving back vowels \cite{Myers-Schulz:2013bh}. Correspondences between articulatory gestures and the shapes of objects have been found already in toddlers, who associate back produced vowels as in `bouba' with rounded forms  whereas bright front vowels  as in `kiki'  are associated with edgy outlines \cite{Maurer:2006aa}. Phonological cues might thus provide a semantic bootstrapping that would facilitate language learning \cite{Monaghan:2007aa}. If aspects of action based language have through Hebbian learning been associated with sequences of verbs and nouns \cite{Glenberg:2011vn}, the underlying parameters of motion could potentially also be retrieved from the latent semantics of action verbs. To explore whether spatiotemporal parameters might also be reflected in the phonetic structure of action verbs, 3 $\times$ 20 emotion, face, and hand related verbs were selected, which have in earlier neuroimaging studies been shown to activate motor, premotor and prefrontal circuits in the brain during a passive reading task \cite{Moseley:2011ys}. Applying latent semantic analysis LSA \cite{Landauer:1997ve} to define the cosine similarities between the action verbs, adjacency matrices were computed based on two different large scale text corpora; HAWIK, consisting of 22829 words found in 67380 excerpts of Harvard Classics literature, Wikipedia articles and Reuters news \cite{Petersen:2012aa} and TASA, consisting of 92409 words extracted from 37651 text excerpts, reflecting the educational material US students have been exposed to when entering their first year of college \cite{Landauer:2007aa}. To identify significant structures among the action verbs, hierarchical clustering was applied to the two adjacency matrices derived from the HAWIK and TASA corpora, using multistep-multiscale bootstrap resampling with Pearson correlation as distance measure \cite{Shimodaira:2004aa}. Selecting the action verbs which were grouped similarly based on both the HAWIK and TASA text corpora, the words were annotated with their corresponding user rated word norms related to the psychological dimensions of valence and arousal \cite{Warriner:2013aa}, thus defining their emotional polarity and perceived intensity \cite{Russell:1980qf}. Subsequently the action verbs were transformed into ARPAbet phonemes using the CMU text to speech pronunciation dictionary \cite{CMU:1976aa}, and their primary stress vowels projected into an articulatory space defined by tongue height and front-back position in the international phonetic alphabet (IPA) and framed by their corresponding F1 and F2 formant frequencies \cite{Catford:1988ys}. \clearpage

\section*{Results and Discussion}

\begin{figure}[ht]
 \centering
  \centerline{\includegraphics[width=12cm]{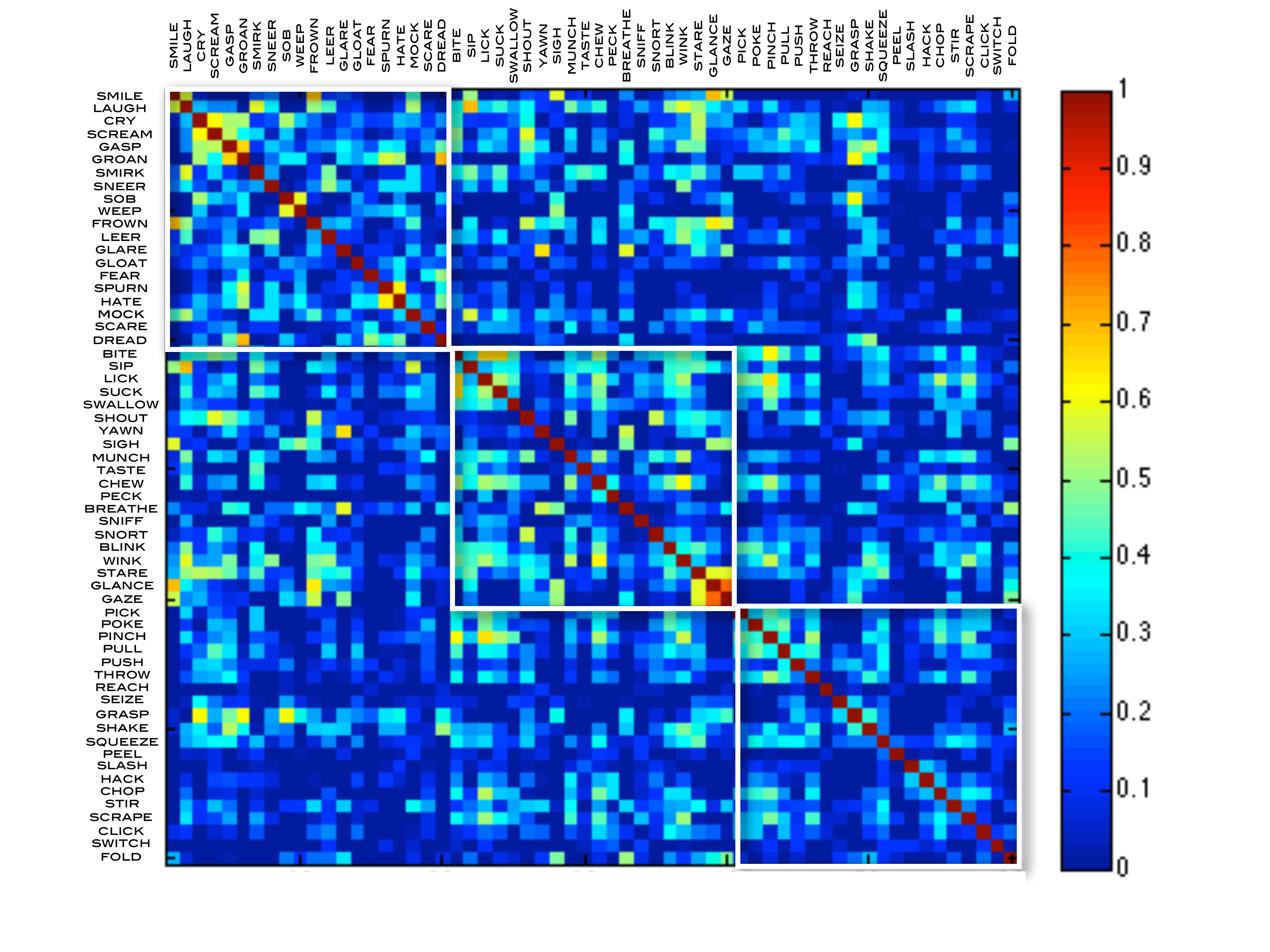}}
  \vspace{2.0cm}
\caption{$ \ \mathbf{HAWIK} \ \mathbf{adjacency} \ \mathbf{matrix} \ \mathbf{of} \ \mathbf{emotion,} \ \mathbf{face} \ \mathbf{and} \ \mathbf{hand}  \  \mathbf{action} \ \mathbf{verbs,}$ defining their cosine similarities generated by applying LSA latent semantic analysis and SVD singular value decomposition to reduce the dimensionality to the 125 most significant eigenvalues. The latent semantic relations between the 3 $\times$ 20 verbs are computed based on the HAWIK text corpus term document co-occurrence matrix, consisting of 22829 words extracted from 67380 excerpts of Harvard Classics literature, Wikipedia articles and Reuters news.}
\label{fig:HAWIKmatrix}

\end{figure}

\begin{figure}[ht]
  \centering
  \centerline{\includegraphics[width=12cm]{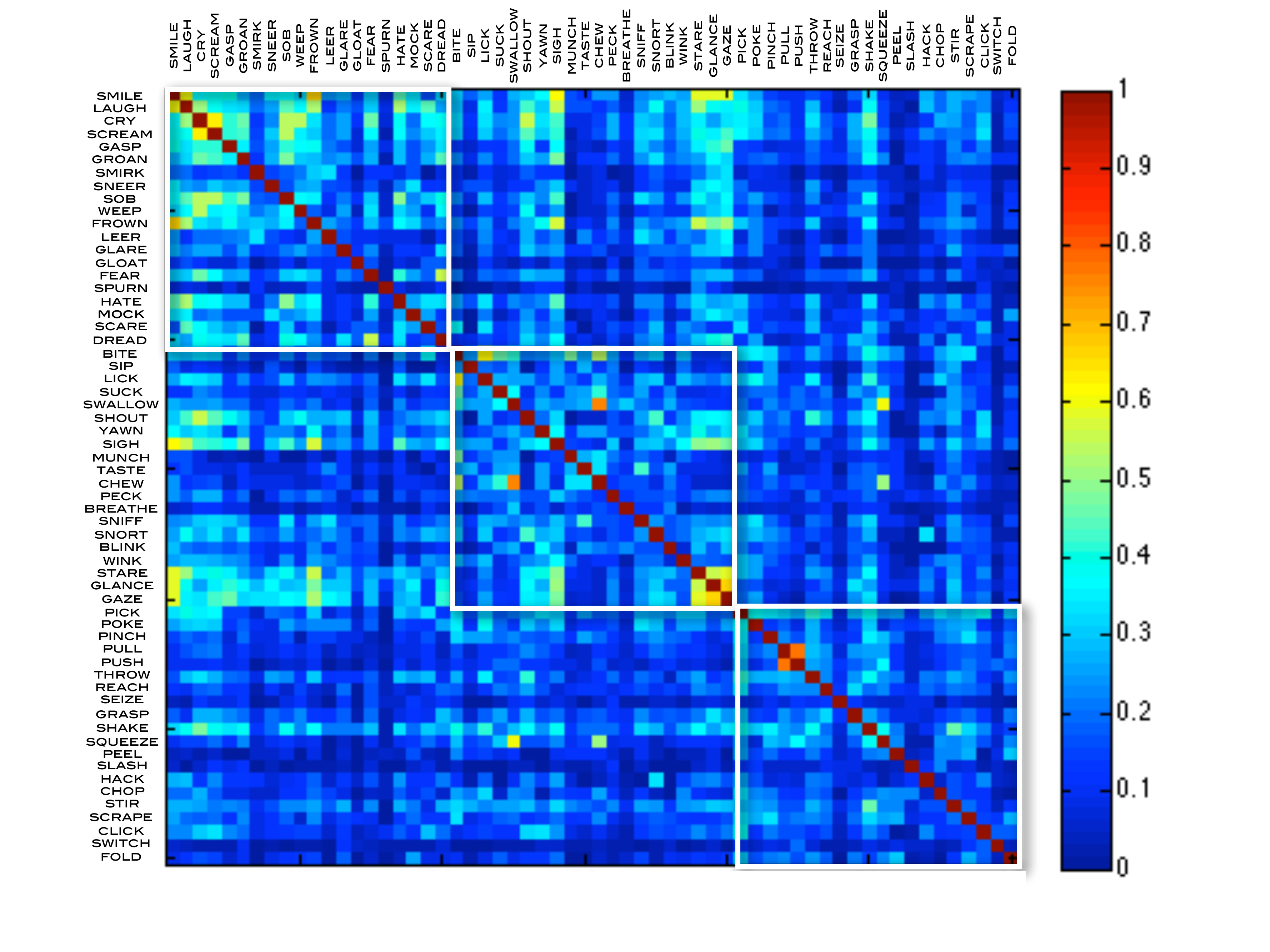}}
  \vspace{2.0cm}
\caption{$ \ \mathbf{TASA} \ \mathbf{adjacency} \ \mathbf{matrix} \ \mathbf{of} \ \mathbf{emotion,} \ \mathbf{face} \ \mathbf{and} \ \mathbf{hand}  \  \mathbf{action} \ \mathbf{verbs,}$ defining their cosine similarities generated by applying LSA latent semantic analysis and SVD singular value decomposition to reduce the dimensionality to the 300 most significant eigenvalues. The latent semantic relations between the 3 $\times$ 20 verbs are computed based on the TASA text corpus term document co-occurrence matrix, consisting of 92409 words extracted from 37651 text excerpts, reflecting the educational material US students have been exposed to when entering their first year of college.}
\label{fig:TASAmatrix}

\end{figure}

\begin{figure}[ht]
  \centering
  \centerline{\includegraphics[width=18cm]{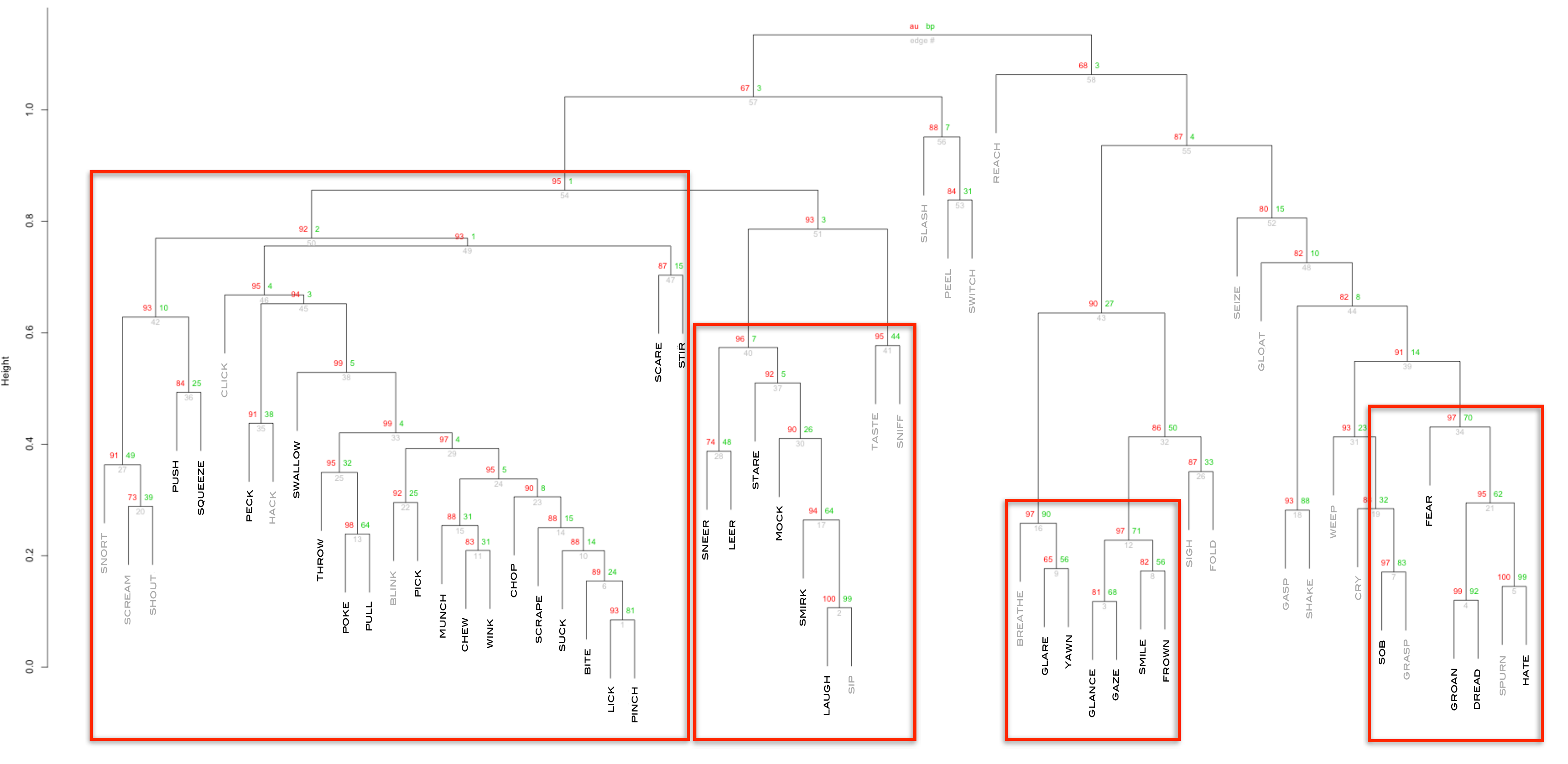}}
  \vspace{2.0cm}
\caption{$\ \mathbf{Hierarchical}  \ \mathbf{clustering} \ \mathbf{of} \  \mathbf{action} \ \mathbf{verbs} \ \mathbf{based} \ \mathbf{on} \ \mathbf{HAWIK}  \  \mathbf{adjacency} \ \mathbf{matrix,}$ using multistep-multiscale bootstrap resampling (n $=$ 100.000) with Pearson correlation as distance measure. AU approximately unbiased  p $-$ values (p $\leq$ 0.05, SD $\leq$  0.004) define significant clusters as red rectangles (bold fonts denoting verbs similarly clustered based on the TASA dendogram) from left to right: 1. Mouth and hand movements of low to high arousal (2.81 - 7.10, M $=$ 4.44)  2. Emotional contrasting low versus high valence  (3.30 - 7.53, M $=$ 4.45)  3. Facial expressions of low arousal (2.63 - 4.52, M $=$ 3.83) 4. Negative emotions of low valence (1.96 - 3.90, M $=$ 2.93)}
\label{fig:HAWIK100Kdendogram}

\end{figure}

\begin{figure}[ht]
  \centering
  \centerline{\includegraphics[width=8cm]{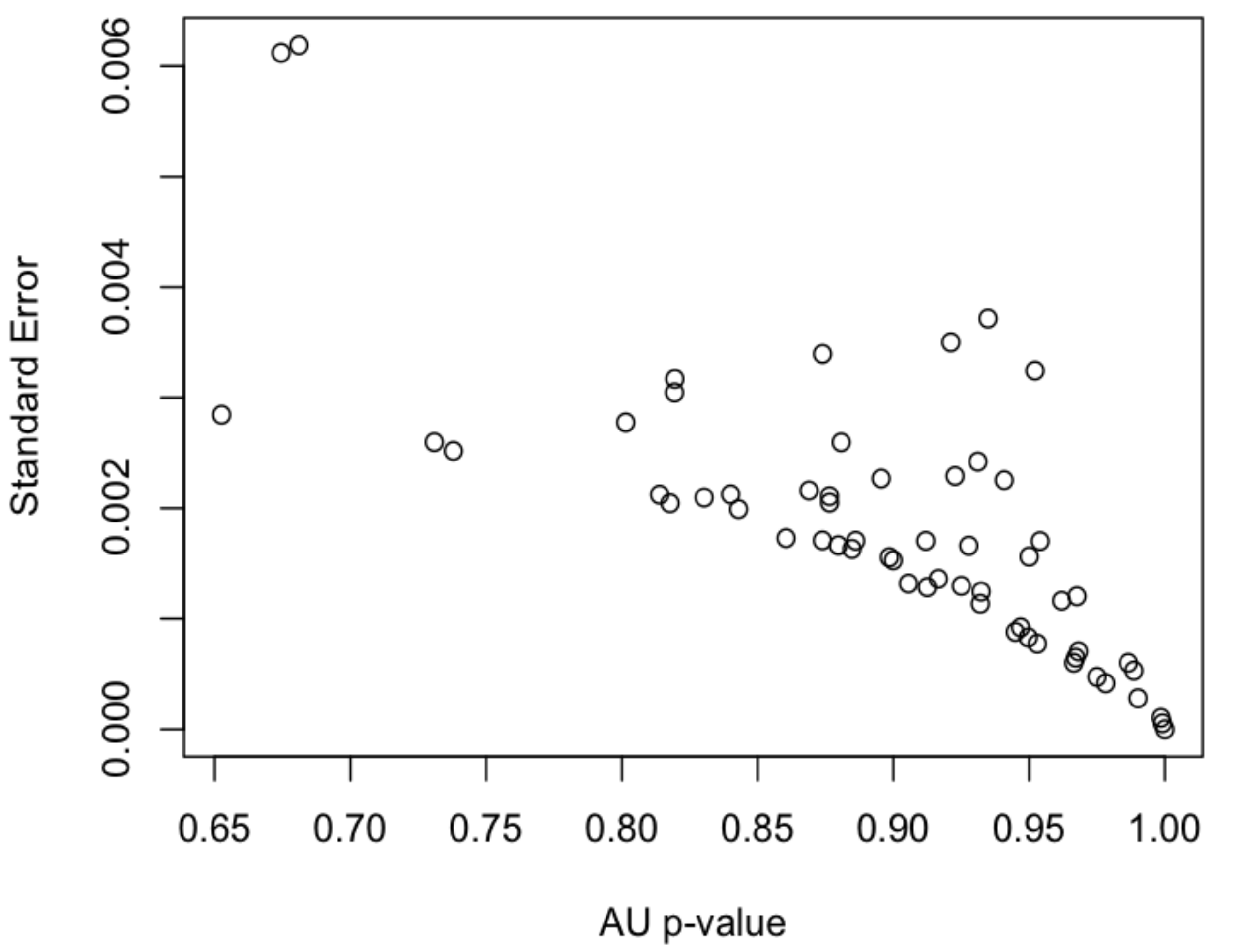}}
  \vspace{2.0cm}
\caption{$\ \mathbf{AU}  \ \mathbf{p-values} \ \mathbf{plotted}  \ \mathbf{against} \ \mathbf{standard} \ \mathbf{errors} $ for the clusters identified in the HAWIK adjacency matrix, computed using multistep-multiscale bootstrap resampling (n $=$ 100.000), indicating standard errors $\leq$  0.004 for the AU approximately unbiased p-values $\leq$ 0.05 used to reject the null hypothesis of the hierarchical clustering structure not being supported by the data.}
\label{fig:HAWIK100Kseplot}

\end{figure}

\begin{figure}[ht]
  \centering
  \centerline{\includegraphics[width=18cm]{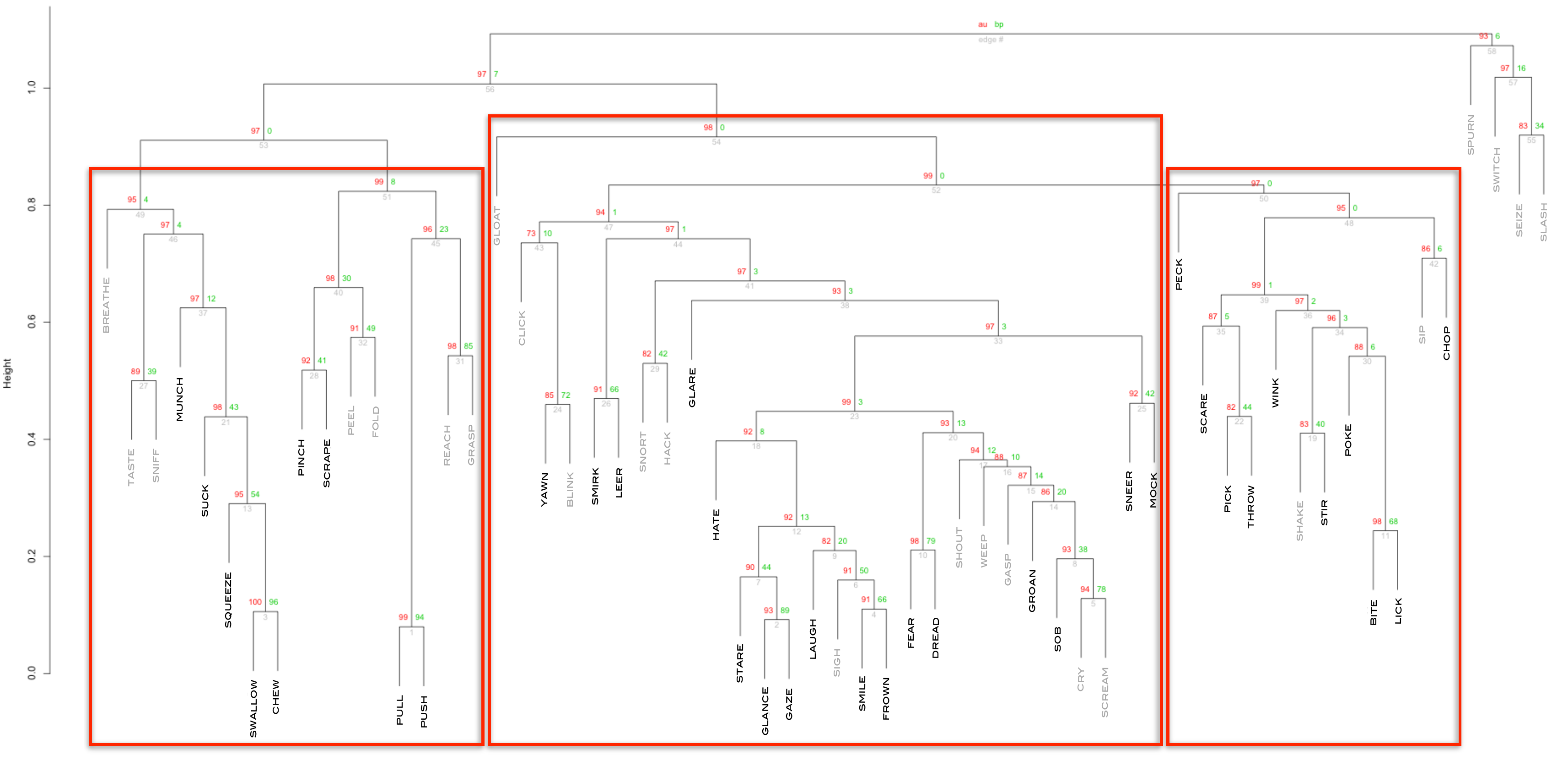}}
  \vspace{2.0cm}
\caption{$\ \mathbf{Hierarchical}  \ \mathbf{clustering} \ \mathbf{of} \  \mathbf{action} \ \mathbf{verbs} \ \mathbf{based} \ \mathbf{on} \ \mathbf{TASA}  \  \mathbf{adjacency} \ \mathbf{matrix,}$ using  multistep-multiscale bootstrap resampling (n $=$ 100.000) based on Pearson correlation with approximately unbiased AU p $-$ values (p $\leq$ 0.05, SD $\leq$  0.007) define significant clusters as red rectangles (bold fonts denoting verbs similarly clustered based on the HAWIK dendogram) from left to right: 1. Mouth and hand movements of low to medium arousal (2.60 - 5.60, M $=$ 3.98) 2. Emotional expressions of low versus high valence (1.96 - 7.89, M $=$ 3.95) 3. Mouth and hand motion characterized by enhanced dominance (4.58 - 6.74, M $=$ 6.12)}
\label{fig:TASA100Kdendogram}

\end{figure}

\begin{figure}[ht]
  \centering
  \centerline{\includegraphics[width=8cm]{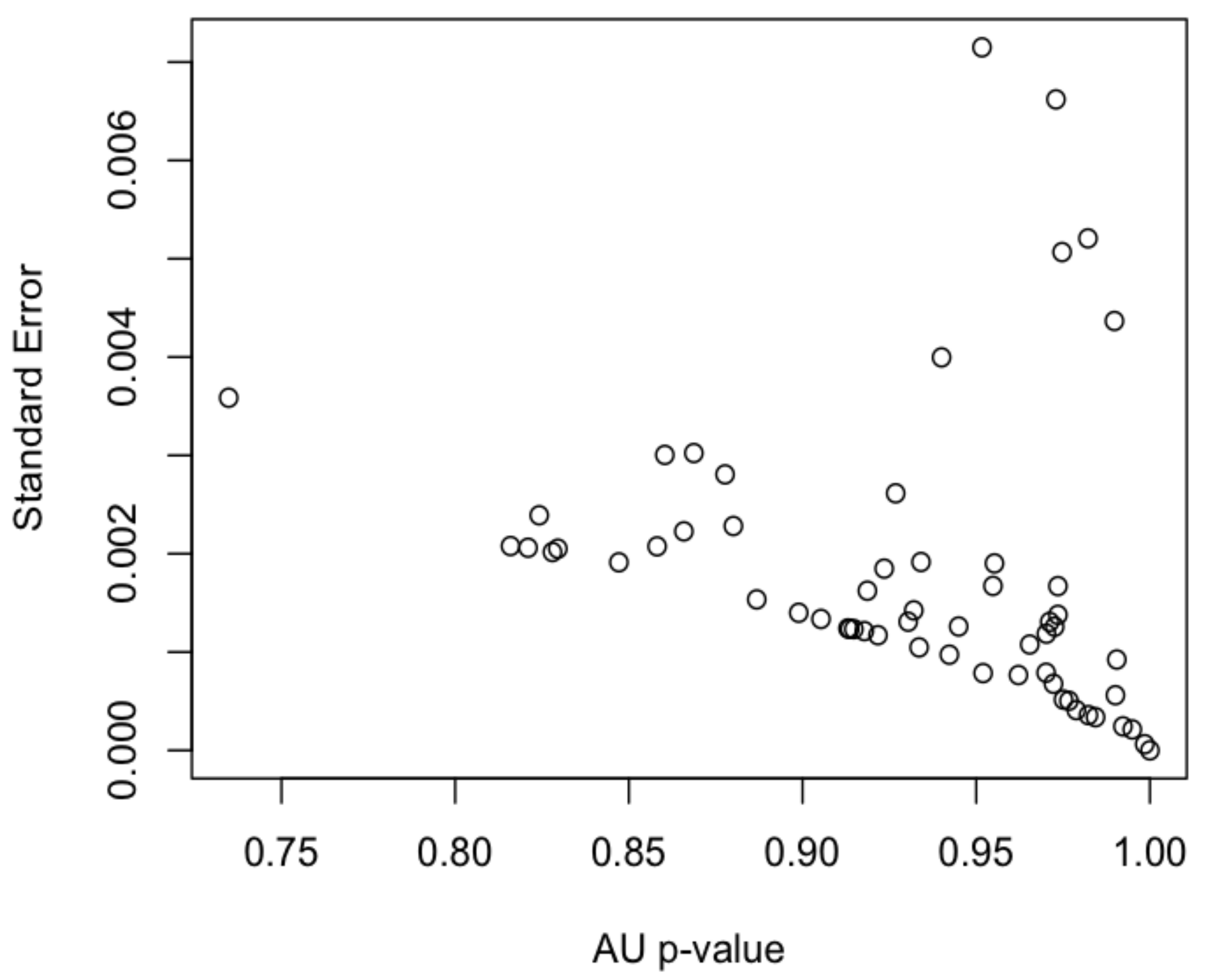}}
  \vspace{2.0cm}
\caption{$\ \mathbf{AU}  \ \mathbf{p-values} \ \mathbf{plotted}  \ \mathbf{against} \ \mathbf{standard} \ \mathbf{errors} $ for the clusters identified in the TASA adjacency matrix, computed using multistep-multiscale bootstrap resampling (n $=$ 100.000), indicating standard errors $\leq$  0.007 for the AU approximately unbiased p-values $\leq$ 0.05 used to reject the null hypothesis of the hierarchical clustering structure not being supported by the data.}
\label{fig:TASA100Kseplot}

\end{figure}

\begin{figure}[ht]
  \centering
  \centerline{\includegraphics[width=12cm]{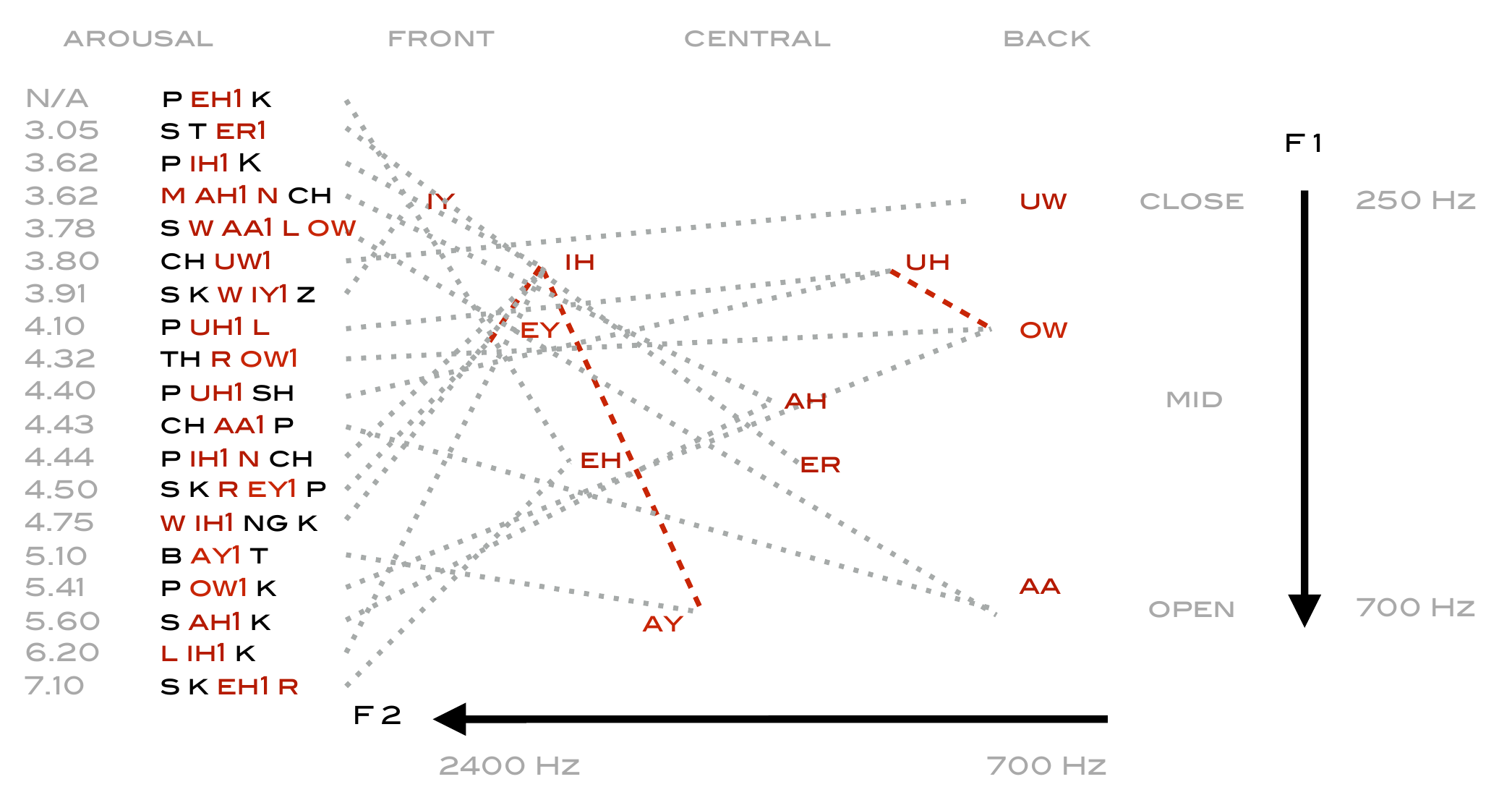}}
  \vspace{2.0cm}
\caption{$\ \mathbf{Articulatory} \ \mathbf{projections}  \ \mathbf{of}  \ \mathbf{primary}  \ \mathbf{stress}  \ \mathbf{vowels} \ \mathbf{in} \ \mathbf{mouth} \ \mathbf{and} \ \mathbf{hand} \ \mathbf{verbs}$ horizontally differentiate front versus back vowels (dotted) and diphthongs (dashed) mapped according to tongue height, front-back position and rounding, whereas auditory features are defined by the corresponding F1 and F2 formant frequencies.}
\label{fig:HAWIKTASAarousal}

\end{figure}

\begin{figure}[ht]
  \centering
  \centerline{\includegraphics[width=12cm]{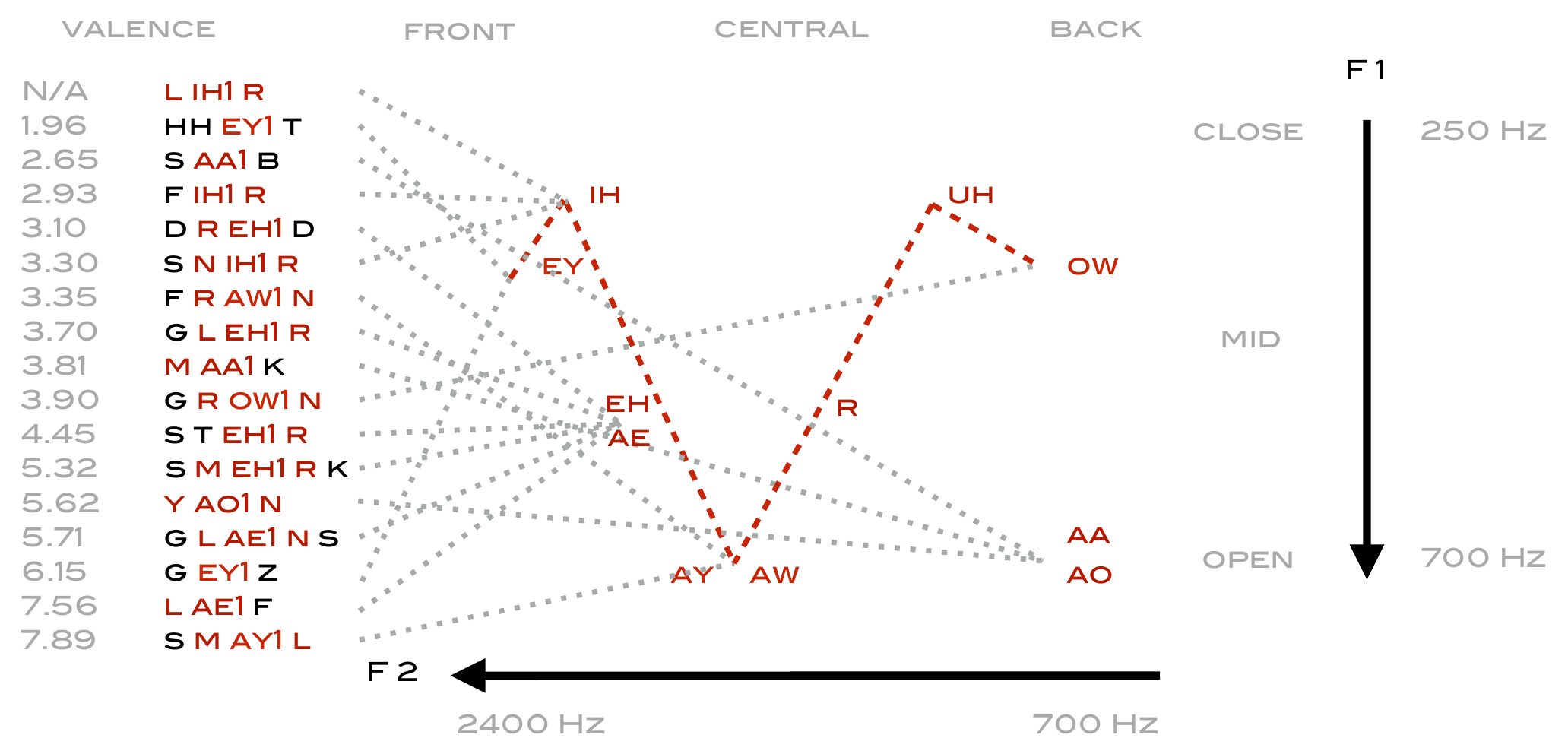}}
  \vspace{2.0cm}
\caption{$\ \mathbf{Articulatory} \ \mathbf{projections}  \ \mathbf{of}  \ \mathbf{primary}  \ \mathbf{stress}  \ \mathbf{vowels} \ \mathbf{in} \ \mathbf{facial} \ \mathbf{and} \ \mathbf{emotional} \ \mathbf{verbs}$ of positive and negative valence, vertically contrast open versus close vowels (dotted) and diphthongs (dashed), mapped according to tongue height, front-back position and rounding, while auditory features are defined by the corresponding F1 and F2 formant frequencies.}
\label{fig:HAWIKTASAvalence}

\end{figure}

\clearpage

\subsection*{Hierarchical structures}

In the LSA adjacency matrices generated using the HAWIK (Fig. \ref{fig:HAWIKmatrix}) and TASA (Fig. \ref{fig:TASAmatrix}) text corpora, the co-occurrences of verbs in the upper left quadrant stand out. Here emotions such as `smile' and `frown' are coupled with facial  expressions like `gaze' and `glance' further down along the diagonal, but only sparsely trigger hand related verbs such as `pick' or `poke' in the lower right corner of the matrix. Whereas verbs related to cyclical jaw and tongue motion such as `bite'  or `chew' strongly co-activate hand movements like `pinch' in the HAWIK or `squeeze' in the TASA corpora.  Behaviorally, combinations of hand and mouth movements  typically co-occur in sequences when we pick up something using the fingers, rotating the elbow and moving the hand towards the face while simultaneously opening the mouth \cite{Graziano:2002aa}. Such sequences of frequently co-occurring movements appear to form a vocabulary of hierarchically structured motor schemas, which are recursively combined into object oriented patterns of motion. In effect reducing the complexity when manipulating objects from a large number of free parameters to only a few dimensions of orientation and size. These parameters might in turn be related to stored representations of sequences constrained by physical parameters of distance and gravity \cite{Jeannerod:1995aa}  \cite{Konkle:2012ly}. 

To quantify the hierarchical structures among the action verbs, characterized by heavily interconnected subgraphs which are only sparsely linked to other modules \cite{Bullmore:2009nx}, hierarchical clustering was applied to the HAWIK and TASA adjacency matrices. Assessing to what degree the resulting hierarchical tree structures are supported by the data,  approximately unbiased AU p-values were computed by multistep multiscale resampling changing the sample size of the bootstrap replicates \cite{Shimodaira:2004aa} to define statistically significant clusters based on the HAWIK  (Fig \ref{fig:HAWIK100Kdendogram} \& \ref{fig:HAWIK100Kseplot}: p $\leq$ 0.05, SD $\leq$  0.004) and TASA (Fig \ref{fig:TASA100Kdendogram} \& \ref{fig:TASA100Kseplot}: p $\leq$ 0.05, SD $\leq$  0.007) dendograms. Visualized in a dendogram, the x axis represents the connectivity between the pairwise most similar leaf nodes and the y axis defines their distances. The resulting tree thus approximates the structural properties of the data by grouping the nodes at specific levels within a hierarchy \cite{Moreno-Dominguez:2014aa}. When comparing the dendograms to identify common structures across the two text corpora, clusters of combined mouth and hand movements are in both the HAWIK and TASA separated from the clusters of facial expressions and emotions. As exemplified by the action verbs in HAWIK cluster 1  describing small size finger precision grip combined with oscillatory jaw and tongue motion such as  `squeeze', `pinch', `scrape', `suck', `chew', `munch' and `swallow', which are similarly grouped in cluster 1 of the TASA dendogram. More forceful movements in HAWIK cluster 1, involving the arm combined with a whole hand grip like `stir' `pick' `throw' `chop' `poke', plus oscillatory motion as in `bite', `lick', `peck' and `wink', are in the TASA dendogram grouped separately in cluster 3.  Indicating, that motor schemas grouping small size hand and mouth movements involved when eating are also reflected in the latent semantics of action verbs across both the TASA and HAWIK text corpora. Likewise, the hierarchical relations between finger precision motion and more forceful whole hand grips involving the arm, are reflected in the hierarchical structures of action verbs juxtaposing small size gestures against larger movements. Suggesting, that hierarchical linguistic structures reflecting aspects of size and intensity might lower the number of free parameters in action verbs. Similar to how ideal antonym pairs like `pull' / `push' or `smile' / `frown'  conceptually reduce dimensionality by sharing multiple semantic features  while crucially differing along a single parameter \cite{Murphy:1993aa}.

\subsection*{Motion and emotion}

Exploring how the clustered action verbs are perceived along a reduced number of psychological dimensions, they were subsequently annotated with crowd sourced user ratings available online from the ``Norms of valence, arousal and dominance for 13915 English lemmas''  data set \cite{Warriner:2013aa}. Similar to the smaller ANEW data set \cite{Bradley:2010aa} the user ratings assess how pleasant, intense and controlled the verbs are described as being on a scale from 1 to 9, along the psychological dimensions of valence, arousal and dominance \cite{Russell:1980qf}. Resulting in a user rated semantic space framed by three parameters: valence defining how pleasant or unpleasant something is perceived on a bipolar axis (1.26-8.48, M$=$ 5.06, SD $=$ 1.68), arousal capturing the level of intensity going from calm to excited (1.60-7.79, M$=$ 4.21, SD $=$ 2.30) and dominance (1.68-7.74, M$=$ 5.18, SD $=$ 2.16) describing the degree of feeling in control. Rather than considering the verbs as individual terms they are thus represented as emotional buoys in a semantic space, related to aspects of approach or avoidance behavior that are psychologically grounded in motivational survival mechanisms \cite{Lang:2010aa} \cite{LeDoux:2012aa}. Combining the user rated annotations on how the words are perceived with the clustering analysis, the  hierarchical structures of mouth and hand motion verbs in the HAWIK cluster 1 appears to a larger degree to capture the contrasts between low and high intensity (arousal 2.81 - 7.10, M $=$ 4.44), than the differences in emotional polarity (valence 3.55 - 6.62, M $=$ 5.37). In line with studies showing that sensorimotor elements remain more significant for the representation of concrete actions \cite{Kousta:2011fk}, and that the strength of a sensory experience is positively correlated with arousal \cite{Warriner:2013aa}. These mouth and hand verbs are in the TASA dendogram divided up between clusters 1 and 3, constituting small and more intense movements as described above, characterized by low to medium arousal (2.60 - 5.60, M $=$ 3.98) and enhanced dominance (4.58 - 6.74, M $=$ 6.12) respectively.

Verbs describing emotions and facial expressions are grouped together in cluster 2 of the TASA dendogram, characterized by contrasts between high and low valence (valence 1.96 - 7.89, M $=$ 3.95, arousal 2.63 - 6.62 M $=$ 4.56). In general abstract concepts have been shown to increasingly rely on affective associations the more abstract they are perceived as being \cite{Kousta:2011fk}. Emotions and facial expressions are in the HAWIK dendogram divided into three groups, where cluster 2  captures contrasts in emotional polarity ranging from the negative verbs `leer', `sneer', `mock' and `stare', to the positive verbs `smirk' and `laugh' (valence 3.30 - 7.56, M $=$ 4.45; arousal 4.36 - 6.62 M $=$ 4.7), while cluster 3 is grouping facial expressions of low intensity in the verbs `frown', `glare', `yawn', `glance', `gaze' and `smile' (arousal 2.63 - 4.52, M $=$ 3.83; valence 3.35 - 7.89, M $=$ 5.7 ).  Thus grouping verbs  characterized by both high valence and low arousal values, which when combined have been found to to strengthen motivational approach behavior \cite{Recio:2014aa}. Whereas the HAWIK cluster 4 isolates the most negative verbs `hate', `sob', `fear', `dread' and `groan' defined by values at the low end of the valence dimension (valence 1.96 - 3.90, M $=$ 2.93; arousal 4.17 - 6.26, M $=$ 4.89). Previous studies have established  that the three dimensions are not orthogonal, as high or low levels of both valence and dominance are rated as more arousing the further away from neutral they are perceived as being \cite{Warriner:2013aa}. That is, the hierarchical clustering of emotional verbs thus seem to separate unpleasant feelings rated low in valence as in the verbs `sob', `fear', `dread' and `groan', from relaxed expressions rated low in arousal like `smile' `frown', `glance' and `gaze'.

\subsection*{Phonetic parameters}

To analyze whether latent semantic parameters of size and intensity might be reflected in the phonetic building blocks of action verbs, their primary stress vowels were spatially mapped out in an articulatory space  according to the international phonetic alphabet IPA, framed by their corresponding acoustical F1 and F2 formant frequencies \cite{Ladefoged:1989fk}. Although speech sounds constitute a multidimensional feature space, we primarily perceive phonemes by the way in which they are articulated and secondarily by where they are produced in the vocal tract. Meaning, that the main contrast is between sonorant vowels versus plosive and fricative obstruents \cite{Mesgarani:2014aa}. In the present study, the hierarchically clustered action verbs describing small size object manipulation appear characterized by high frontal vowels as in `P IH1 K', `P IH1 N CH', `S K W IY1 Z', `W IH1 NGK' and `L IH1 K', while more forceful  actions required for moving or transforming objects incorporate phonemes produced further back as in `P UH1 L' , `P UH1 SH', `P OW1 K', `CH UW1', `CH AA1 P', `S W AA1 L OW' and `TH R OW1'  (Fig. \ref{fig:HAWIKTASAarousal}). That is, the small size gestures are acoustically defined by high F2  formant frequencies that are maximally dispersed from the F1 formants. While more forceful actions are articulated by back vowels and diphthongs which acoustically have a smaller gap between the F2 and F1 formant frequencies. 

In articulatory terms, as the tongue is gradually raised towards the palate, the sonorants undergo a phase shift when the airflow turns turbulent and vowels are transformed into fricatives and plosive stops. Among the action verbs in this study, small movements such as `P IH1 K' and `P EH1 K' are characterized by both plosive attacks as well as stops which abruptly cut off the resonance of the sonorants. More forceful movements incorporate affricates which begin as plosives and release as fricatives like in `CH UW1' or `CH AA1 P'. Whereas verbs initiated by fricatives create a feeling of sustained tension by forcing the flow of air over the edge of the teeth as in  `S K R EY1 P' and `S AH1 K'. Likewise aspects of contact as in `S K W IY1 Z' and `P UH1 SH' are emphasized by fricatives sustaining the airflow towards the end of the verb. It has been suggested that the phonetic building blocks of plosives, sonorants, and fricatives may have evolved by mimicking the sounds that occur when physical objects collide, resonate or slide across a surface. Interpreted in that sense phonemes may encapsulate additional perceptual characteristics, as verbs beginning with an unvoiced `P' as in `P OW1 K' temporally extends the gap before the diphthong `OW', thereby creating a resonance that acoustically resembles the impact of soft objects with a flexible texture. Whereas the gap before the sonorant is shortened when applying a voiced plosive `B' as in `B AY1 T', resulting in a resonance that acoustically would be associated with collisions of larger more rigid structures \cite{Changizi:2011zr}. The parameters defined by the dispersion between the F2 and F1 formant frequencies may additionally provide emotional cues that differentiate the hierarchically clustered facial expressions and emotional verbs (Fig. \ref{fig:HAWIKTASAvalence}). Dynamically lowering the pitch in vowels is perceived as threatening in human speech sounds, while upwards moving formant transitions are to a larger degree associated with positive emotions \cite{Myers-Schulz:2013bh}. Such up- or downward shifts in pitch of the F2 formants are evident in the diphthong double vowels of `S M AY1 L' and `G EY1 Z' moving towards the front `IH' or in `F R AW1 N' towards the back `UH'. These contrasts are amplified by transforming the white noise-like fricatives `S', `F' or `H'  into alternating closed and open jaw produced  sequences of phonemes as in `S AA1 B', `F R AW1 N' and `H EY1 T'. Earlier studies have based on latent semantics established that phonaesthemes such as the prefix gl- frequently occurring in verbs like `G L AE1 N S', are not arbitrary phonemes but capture conceptual patterns related to vision \cite{Otis:2008aa}, that can be traced back to common linguistic Proto-Indo-European roots \cite{Boussidan:2009aa}. Such visual semantic associations are phonetically defined by consonants rather than vowels, but the upwards or downwards formant shifts in `G EY1 Z'  or `G L EH1 R', suggests that additional emotional content could be encoded in the phoneme transitions. Likewise, downward frequency shifts due to the lowered F3 third formant characteristic of the liquid consonant `R', are in this study prevalent in negative emotions such as `L IH1 R', `F IH1 R', `S N IH1 R', `G L EH1 R', `S T EH1 R' and `S K EH1 R', that are all rated below 4.50 in perceived valence.

Whether language is seen as grounded in simulation literally dependent on sensorimotor circuits, or rooted in symbolic associations constituted by statistical word representations, there is an emerging consensus on the need to adapt a pluralist view about embodiment and semantics \cite{Meteyard:2012aa} \cite{Pulvermuller:2013fk}. Spatial parameters have in previous studies been extracted from sentences describing horizontal or vertical movements based on latent semantics retrieved from word co-occurrences \cite{Louwerse:2011aa}. Potentially based on sensorimotor links between perception of shapes and motion, which may have constrained how aspects of size and intensity are mapped onto the consonants and vowels of words \cite{Ramachandran:2001aa}. Underlying structural dimensions of size seem hardwired into speech articulation, as grasping objects of increasing size has been shown to simultaneously enlarge both the lip kinematics and mouth aperture when pronouncing vowels \cite{Gentilucci:2006fk}. The phonetic contrasts differentiating the verbs in the present study are in line with recent findings related to how phonetic features re represented in the brain when listening to speech, showing that the relation between the F2 and F1 formants constitutes the first principal component in neuronal encoding of vowels. Indicating that a higher order encoding of acoustic formant parameters provide the foundation for neuronal response properties that are maximally tuned to differentiate open low back from close high front vowels \cite{Mesgarani:2014aa}. In the present study exemplified by the contrasts between the primary stress vowels in `CH AA1 P' versus `P IH1 K'. Likewise when pronouncing consonant vowel syllables, the hierarchical structure of consonants primarily reflects whether they are produced by the lips or dorsal tongue as in `P' or `K' versus coronal tongue as in `CH' or `SH'. While the hierarchical structure of vowels at the highest level differentiate unrounded high front and low back vowels as in `P IH1 K' and `CH AA1 P' from rounded high back vowels as in `P UH1 SH' \cite{Bouchard:2013aa}. These articulatory structures appear in the present study reflected in the latent semantic contrasts between small size motion as in `P EH1 K' , combining labial and dorsal consonants with unrounded vowel, versus forceful movements such as `TH R OW1' combining coronal consonant with rounded vowel. A simplified representation of phonetic features could be interpreted as a continuum, vertically going from open jaw low back vowels such as `AA'  ` in `S W AA1 L OW' to near close frontal vowels such as `IY' in `S K W IY1 Z',  that are in turn horizontally contrasted against high back rounded vowels like `UH'  in `P UH1 L'.  Suggesting, that these articulatory and acoustic dimensions not only maximize phonetic contrasts to facilitate comprehension, but also reflect spatial parameters of intensity and emotional content encoded in action verbs, that can be extracted from word co-occurrences in the surface structure of language.

\section*{Methods}

Initially $3 \times 20$ hand, face and emotion related action verbs were selected, constituting half of the action verbs similarly used in a fMRI functional magnetic resonance neuroimaging study, demonstrating that the selected action verbs activated premotor cortices in the brain during a passive reading task \cite{Moseley:2011ys}. Latent semantic analysis LSA \cite{Deerwester:1990ys} \cite{Landauer:1997ve} was applied in order to retrieve an adjacency matrix based on the HAWIK text corpus consisting of 22829 words found in 67380 excerpts of Harvard Classics literature, Wikipedia articles and Reuters news \cite{Petersen:2012aa}. The cosine similarities of verbs based on the HAWIK text corpora were subsequently compared against those retrieved based on the  TASA corpus consisting of 92409 words in 37651 fiction and non-fiction texts extracted from novels, news articles, and other general knowledge reading material that the average American student has been exposed to from 3rd grade until reaching first year of college \cite{Landauer:2007aa}. Using singular value decomposition SVD to reduce dimensionality \cite{Furnas:1988oq}, the original $m \times  n$ term-document matrix $\mathbf{X}$ is decomposed into a product of three other matrices:\newline

$\mathbf{X} =  \mathbf{U} \mathbf{\Lambda} \mathbf{V^{T}}$ \newline

\noindent where the $\mathbf{U}$ matrix, similar to the original matrix has $m$ rows of words, while the columns now consist of $r$ eigenvectors representing the principal components in the data. Likewise the transpose of the orthonormal matrix $\mathbf{V^{T }}$ has as before $n$ columns of documents but now related to $r$ rows of eigenvectors or principal components. The very purpose of the decomposition is to scale down the number of parameters based on a $\mathbf{\Lambda}$ square matrix containing $r$ singular values $\lambda$ arranged along the diagonal in decreasing order, which as eigenvalues scale the eigenvectors of the rectangular matrices to each other and thereby derive a matrix of reduced dimensionality:\newline

$\mathbf{Z_{k}} =  \mathbf{U_{k}} \mathbf{\Lambda_{k}} \mathbf{V_{k}^{T}}$ \newline

\noindent where only the $k$ largest singular values of the $\mathbf{\Lambda}$ diagonal matrix are retained. As a result the number of parameters in the rectangular $\mathbf{U_{k}}$ and $\mathbf{V_{k}^{T}}$ matrices are reduced to what would correspond to the principal components containing the highest amount of variance in the matrix. Thus allowing us to reconstruct the original input based on a $\mathbf{Z_{k}}$ matrix of lower dimensionality which is embedding the underlying structure of the data. Geometrically speaking, the terms and documents in the condensed $\mathbf{Z_{k}}$  matrix can be interpreted as points in a $k$ dimensional subspace, which enables us to calculate the degree of similarity between matrices based on the dot or inner product of their corresponding vectors. Interpreting the matrix multiplication geometrically the cosine similarity between two words represented by their vectors can be expressed as \newline

$\displaystyle \cos\theta = \frac{x \cdot y} {\Vert x\Vert \Vert y \Vert}$\newline

\noindent where $x \cdot y$ signifies the dot product of the vectors, and $\Vert x\Vert \Vert y \Vert$ the Euclidean norm corresponding to the square root of the dot product of each vector with itself. \newline 

\noindent The optimal number of 300 dimensions used for the LSA analysis based on the TASA has been determined based on a synonymy test \cite{Landauer:1997ve}. To determine the optimal number of dimensions for the HAWIK corpus a similar synonymy test was implemented, which based on questions from the TOEFL `test of english as a foreign language' compared the LSA cosine similarity of the multiple choice test synonyms, while varying the number of eigenvectors until an optimal percentage of correct answers were returned.  For the HAWIK matrix the optimal result of 71,2\% correctly identified synonyms based on LSA was found when reducing the singular value decomposition SVD to the most significant 125 eigenvalues. This result is above the 64.5\% TOEFL average test score achieved by non-native speaking US college applicants, on par with previous results obtained using either LSA or probabilistic topic models \cite{Griffiths:2007ly}. 

To identify significant structures among the action verbs, hierarchical clustering was applied to the two adjacency matrices derived from the HAWIK and TASA corpora, using Pearson correlation as distance measure. Assessing to what degree the resulting hierarchical tree structures are supported by the data,  approximately unbiased AU p-values were computed by multistep multiscale resampling changing the sample size of the bootstrap replicates \cite{Shimodaira:2004aa}. Selecting the action verbs which were hierarchically clustered similarly based on both the HAWIK and TASA adjacency matrices, the words were annotated with their corresponding user rated word norms available online from the ``Norms of valence, arousal and dominance for 13915 English lemmas''  data set \cite{Warriner:2013aa}. Similar to the smaller ANEW data set \cite{Bradley:2010aa} the user ratings assess how pleasant, intense and controlled the verbs are described as being on a scale from 1 to 9, along the psychological dimensions of valence, arousal and dominance \cite{Russell:1980qf}.

Subsequently the action verbs were transformed into ARPAbet phonemes using the CMU text to speech pronunciation dictionary \cite{CMU:1976aa}, and their primary stress vowels projected into an articulatory space defined by tongue height and front-back position in the international phonetic alphabet (IPA), to identify common spatial parameters acoustically framed by the corresponding average F1 and F2 formant frequencies \cite{Catford:1988ys}.


\bibliography{refs}

\end{document}